\documentclass{article}

\PassOptionsToPackage{numbers, compress}{natbib}

\usepackage[preprint]{neurips_2025}

\usepackage[utf8]{inputenc} 
\usepackage[T1]{fontenc}    
\usepackage{hyperref}       
\usepackage{url}            
\usepackage{booktabs}       
\usepackage{amsfonts}       
\usepackage{nicefrac}       
\usepackage{microtype}      
\usepackage[table]{xcolor}         
\usepackage{graphicx}
\usepackage{multirow}
\usepackage{listings}
\usepackage{helvet}
\usepackage{amsmath}
\usepackage{algorithm}
\usepackage{algpseudocode}
\usepackage{amssymb}

\usepackage[breakable]{tcolorbox}

\title{AutoLayout: Closed-Loop Layout Synthesis via Slow-Fast Collaborative Reasoning}

\author{%
  Weixing Chen$^{1,2}$ \hspace{0.4em} Dafeng Chi$^{2}$ \hspace{0.4em} Yang Liu$^{1,4}$ \hspace{0.4em} Yuxi Yang$^1$ \hspace{0.4em} \textbf{Yexin Zhang}$^1$ \\ \textbf{Yuzheng Zhuang}$^{2}$ \hspace{0.4em} \textbf{Xingyue Quan}$^{2}$ \hspace{0.4em} \textbf{Jianye Hao}$^{2}$ \hspace{0.4em} \textbf{Guanbin Li}$^{1,4}$ \hspace{0.4em} \textbf{Liang Lin}$^{1,3,4}$\\
  $^1$Sun Yat-sen University, China \hspace{0.4em} 
  $^2$Huawei Noah’s Ark Lab \hspace{0.4em} 
  $^3$Peng Cheng Laboratory\\
  $^4$Guangdong Key Laboratory of Big Data Analysis and Processing\\
}

\begin{document}

\maketitle

\begin{abstract}
The automated generation of layouts is vital for embodied intelligence and autonomous systems, supporting applications from virtual environment construction to home robot deployment. Current approaches, however, suffer from spatial hallucination and struggle with balancing semantic fidelity and physical plausibility, often producing layouts with deficits such as floating or overlapping objects and misaligned stacking relation. In this paper, we propose AutoLayout, a fully automated method that integrates a closed-loop self-validation process within a dual-system framework. Specifically, a slow system harnesses detailed reasoning with a Reasoning-Reflection-Generation (RRG) pipeline to extract object attributes and spatial constraints. Then, a fast system generates discrete coordinate sets and a topological relation set that are jointly validated. To mitigate the limitations of handcrafted rules, we further introduce an LLM-based Adaptive Relation Library (ARL) for generating and evaluating layouts. Through the implementation of Slow-Fast Collaborative Reasoning, the AutoLayout efficiently generates layouts after thorough deliberation, effectively mitigating spatial hallucination. Its self-validation mechanism establishes a closed-loop process that iteratively corrects potential errors, achieving a balance between physical stability and semantic consistency. The effectiveness of AutoLayout was validated across 8 distinct scenarios, where it demonstrated a significant 10.1\% improvement over SOTA methods in terms of physical plausibility, semantic consistency, and functional completeness.
\end{abstract}



\section{Introduction}

The automated generation of layouts is pivotal for advancing embodied intelligence and autonomous systems. Not only does it underpin the construction of virtual environments and enable effective home service robot planning, but it also serves as a critical benchmark for evaluating the spatial reasoning capabilities of vision-language models (VLMs) and large language models (LLMs)~\cite{li2020attribute, gupta2021layouttransformer, zhu2021hierarchical,liu2025aligning}. 
Layout generation is typically required to accommodate imprecise instructions, necessitating complex spatial reasoning aligned with user preferences while ensuring that objects are positioned both efficiently and safely~\cite{kant2022housekeep, xu2024set, Song_2025_CVPR,Luo_2025_CVPR}.
This dual requirement—faithfulness to semantic intent and adherence to the principles of physical stability and structural coherence—presents a formidable challenge~\cite{sun2024layoutvlm}.


\begin{figure}[h]
    \centering
    \includegraphics[width=1\linewidth]{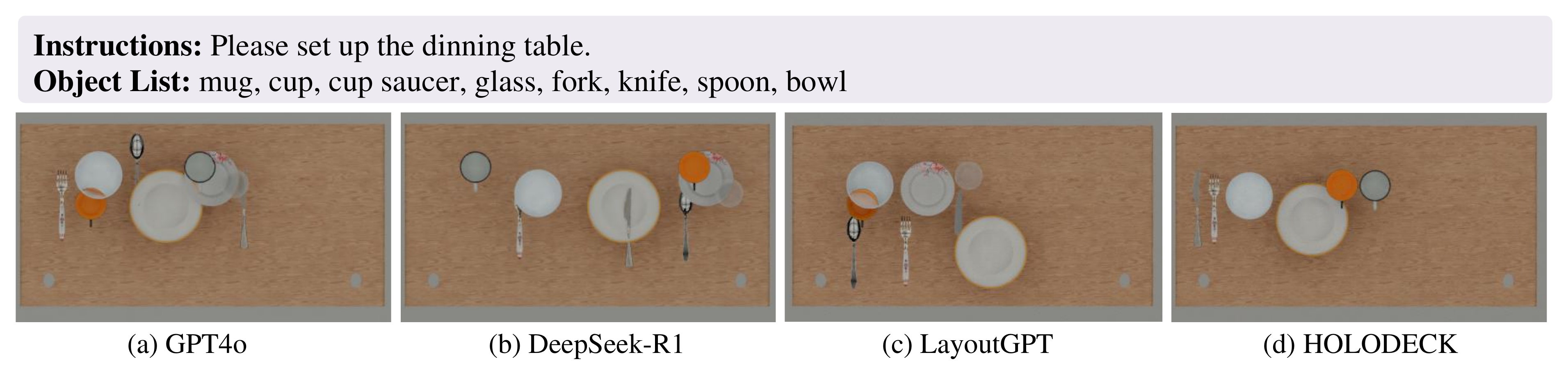}
    \vspace{-10pt} 
    \caption{In the task of generating table layouts, (a-b) directly infer the layouts using the GPT-4o and DeepSeek-R1 models, respectively, with instructions, object lists, and asset dimensions as inputs. (c) employs the LayoutGPT approach, generating layouts with in-context exemplars. (d) The Holodeck approach is the two-stage approach transitioning from coarse to fine granularity. }
    \label{fig:intro_bad_case}
    \vspace{-10pt}
\end{figure}

Despite recent advances, large-scale models continue to struggle with spatial layout reasoning. Both VLMs and LLMs are prone to spatial hallucination~\cite{cheng2024spatialrgpt,Chen_2025_CVPR}, including generating physically infeasible or semantically inconsistent layouts~\cite{liu2025spatialcot}, as shown in Figure \ref{fig:intro_bad_case} (a-b). Moreover, these models often falter when handling fine-grained spatial constraints, especially in tasks demanding precise coordinate positioning or reasoning about complex, multi-layered stacking relation or rearrangement in a limited space~\cite{chen2018iterative}. These shortcomings largely stem from the task’s inherent complexity, which requires the simultaneous integration of abstract semantics, strict geometric constraints, and nuanced physical commonsense capabilities that surpass the typical scope of pretraining scenarios~\cite{bapst2019structured, liu2018physical}.

Existing approaches have attempted to mitigate these issues by adopting an agent-based, two-stage generation process. These methods generate the semantic relations among objects, and the specific coordinates are subsequently derived from these relations \cite{yang2024holodeck, xu2024set}. 
While this strategy improves the physical plausibility of the layouts compared to single-stage approaches \cite{feng2023layoutgpt}, the reliance on semantic relations often results in a neglect of broader contextual information. 
Consequently, the resulting layouts, although physically feasible, may suffer from semantic inconsistencies, as shown in Figure \ref{fig:intro_bad_case} (c-d). 
Furthermore, methods that depend on handcrafted rules or procedural sketches can be effective only within narrowly defined scenarios, as they lack the automation and flexibility required for broader applications.  
To mitigate these issues, models such as LayoutVLM \cite{sun2024layoutvlm} incorporate visual prompts and replace fixed rules with distance and angle based descriptions, they still encounter difficulties in complex stacking tasks. In many cases, objects that complicate spatial reasoning are omitted, which undermines the functional completeness of the layout, as shown in Figure \ref{fig:intro_bad_case} (d).


To address the inherent hallucination issues of LLMs, slow-fast collaborative reasoning was employed to ensure reasoning quality while maintaining layout generation efficiency. Furthermore, a closed-loop process was integrated to continuously self-evaluate layout quality, achieving a harmonious balance among semantic fidelity, physical plausibility, and visual aesthetics.
In our work, we propose AutoLayout, a fully automated layout generation method that leverages a self-consistency validation mechanism alongside a LLM-based Adaptive Relation Library (ARL), all orchestrated by a slow-fast system framework. 

Specifically, the "slow system" creates a comprehensive scene description through an reasoning-reflection-generation (RRG) pipeline, utilizing the ARL created and maintained by LLMs during the layout generation process. This description is then passed to the "fast system" to produce a set of topological relations and discrete coordinate sets. Through self-consistency validation, only those sets meeting strict standards of physical plausibility and semantic consistency are retained.
To ensure functional completeness, an additional validation is performed to identify and address potential errors or omissions, iteratively refining the layout. Subsequently, the discrete coordinate set is merged with object attributes to initialize layout coordinates, while the topological relation set guides the optimization of these coordinates using an evolutionary algorithm. Finally, the alignment of the layout with expected semantic constraints is assessed through a validation function within the ARL. This process, incorporating three stages of self-validation, produces a robust and refined final solution.
Our contributions can be summarized as follows:

\begin{itemize}
    \item To mitigate spatial hallucinations in LLMs/VLMs, we proposed a layout generation platform based on the slow-fast system named AutoLayout. Through meticulous reasoning and efficient generation, the platform ensures that the synthesized outputs meet rigorous physical and semantic standards.
    \item We utilize an LLM-based Adaptive Relation Library, which alleviates the dependency on fixed rules and enhances the method’s applicability across diverse scenarios.
    \item To ensure the functional completeness of the layout, a closed-loop workflow was constructed. Multiple self-validation mechanisms were incorporated to further enhance the reliability of the proposed method.

\end{itemize}

 \section{Related Works}

\subsection{LLMs \& VLMs for 3D Reasoning}
In recent years, LLMs/VLMs have been increasingly explored for their capabilities and limitations in spatial reasoning and spatial hallucinations—issues such as floating, overlapping, or misaligned objects in generated layouts. Works like SpatialRGPT \cite{cheng2024spatialrgpt} enhance local 3D reasoning with regional representations and depth information, while SpatialCoT \cite{liu2025spatialcot} improves multi-step spatial reasoning using bidirectional coordinate alignment and chain-of-thought methods. Similarly, SpatialVLM \cite{chen2024spatialvlm} leverages large-scale 3D question-answering datasets to enable quantitative spatial understanding. However, spatial hallucinations remain a major challenge, as models struggle to balance physical stability and semantic consistency while failing to capture fine-grained spatial constraints in complex environments. Efforts such as EmbodiedVSR \cite{zhang2025embodiedvsr} with dynamic scene graphs and chain-of-thought reasoning, and VoT, which generates internal visual states, have yet to fully resolve these issues~\cite{wu2024mind}. In 3D modeling, CAD-GPT \cite{wang2025cad} maps 3D parameters to linguistic descriptions for precise modeling, while studies like \cite{wang2024can, yang2024thinking} highlight persistent deficiencies in spatial constraints and physical stability. 
To address these limitations, AutoLayout integrates a closed-loop self-validation, combining detailed reasoning and efficient generation to effectively mitigate spatial hallucinations in layout tasks.

\subsection{Layout Generation \& Functional Placement}
In recent years, layout generation and functional object placement have received significant attention as critical challenges in constructing realistic virtual environments and supporting autonomous intelligent systems. Existing methods, while leveraging large language models (LLMs) for semantic planning~\cite{feng2023layoutgpt, yang2024holodeck, sun2024layoutvlm, ccelen2024design}, have achieved notable progress in converting textual descriptions into 2D/3D layouts. 
However, a persistent challenge remains in balancing semantic consistency with physical feasibility. For instance, some approaches focus primarily on semantic guidance, often resulting in layouts plagued by physical flaws, such as floating objects, overlaps, or incorrect stacking relations. 
Other methods attempt to ensure physical stability through hierarchical structures or differentiable optimization but frequently rely on predefined rules or manually designed relational graphs, limiting their ability to support complex human-scene interactions~\cite{luo2020end, gao2023scenehgn, wang2019planit}.

Meanwhile, research on functional object placement has emphasized generating scenes that meet practical human-use requirements~\cite{xu2024set,kant2022housekeep,wei2025functional,li2024llm}. Nevertheless, these approaches often struggle to reconcile semantic and physical constraints in global layout generation and may introduce inconsistencies during local optimization. 
To address these limitations, our method incorporates a adaptive relation library and a slow-fast system framework. This design achieves an effective balance between semantic precision and physical stability, providing a novel and automated solution for layout generation and functional object placement.

\section{AutoLayout}

\begin{figure}
    \centering
    \includegraphics[width=1\linewidth]{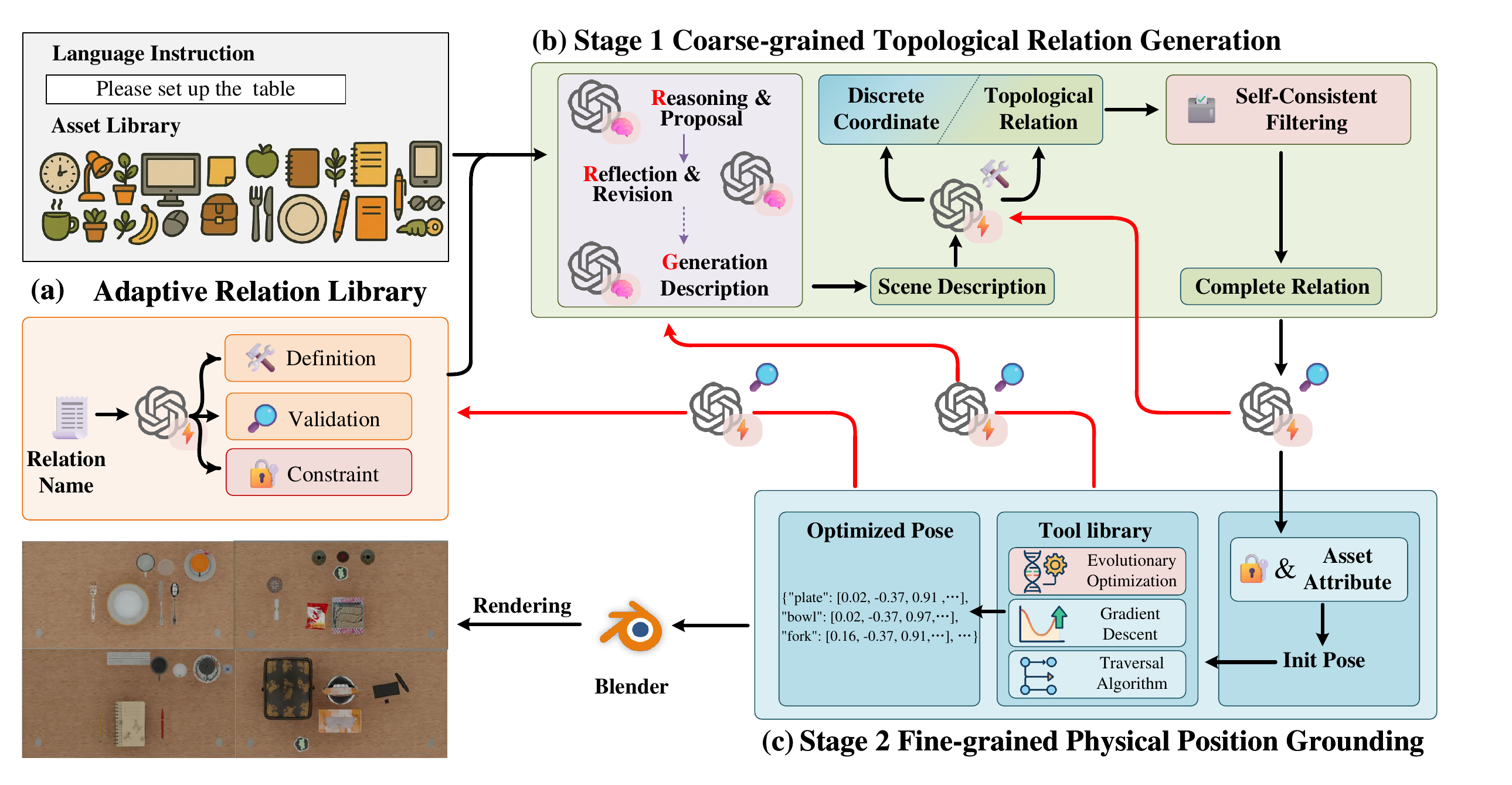}
    \vspace{-15pt}
    \caption{
    The system is driven collaboratively by two subsystems operating at different ``thinking speeds." 
    The slow system (\includegraphics[height=1em]{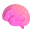}) performs a Reasoning-Reflection-Generation (RRG) chain of reasoning, responsible for extracting and describing high-level semantics.
    The fast system (\includegraphics[height=1em]{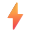}) processes the provided descriptions to rapidly generate or optimize relations and coordinates. It outputs candidate layouts while performing self-validation and adjusting the constraints of the ARL. The red line indicates the self-validation that forms a closed loop.
    }
    \vspace{-10pt}
    \label{fig:method_overview}
\end{figure}

The proposed AutoLayout method is designed within a two-stage, slow-fast system framework to generate physically plausible and semantically compliant layouts. The overall process is divided into two main stages: the first stage focuses on generating coarse-grained topological relations, while the second stage refines the physical positioning, as shown in Figure~\ref{fig:method_overview}. A self-validation mechanism is integrated to iteratively adjust the generated results. The design and implementation details of each module are described below. The symbols are summarized in the Appendix.

Given the following inputs:
A \textbf{language instruction} $\mathcal{I}$ (e.g., "Set up the table"), and
a set of \textbf{objects} $\mathcal{O} = \{O_1, \dots, O_n\}$, where each object is associated with a size vector $\mathbf{s}_i \in \mathbb{R}^3$.
The boundary of the supporting surface $\mathcal{B}$ (e.g., a tabletop or room floor), the goal is to determine a set of coordinates in the 3D space $\Omega \subset \mathbb{R}^3$:

\begin{equation}
    C=\{c_1,...,c_n\}, c_i=[x_i,y_i,z_i,roll_i,pitch_i,yaw_i]
\end{equation}

such that the following conditions are satisfied:

\textbf{Physical Feasibility}:
No object overlap: $\forall i \neq j ; \text{NonOverlap}(O_i, O_j) = \texttt{true}$,
Stability: $\forall i ; \text{Stable}(O_i) = \texttt{true}$.

\textbf{Semantic Consistency}:
$\forall r_k \in \mathcal{R} ; \text{Validation}(r_k, \mathbf{C}) = \texttt{true}$, where $\mathcal{R}$ represents the set of topological relations.

\textbf{Functional Completeness}:
All objects $O_i \in \mathcal{O}$ are placed within the boundary $\mathcal{B}$.

\subsection{Adaptive Relation Library}
The Adaptive Relation Library (ARL) serves as a critical supporting module of the system, enabling dynamic modeling and real-time adjustment of various topological relations, as shown in Figure \ref{fig:method_overview} (a). Traditional methods often rely on fixed, manually crafted rules. In contrast, our system defines a unified I/O format, allowing LLMs to automatically adjust constraint functions when difficulties arise during the solving process. The relation library primarily includes the following three types of relations, shown in Table~\ref{tab:method_relationship}: 

\begin{table}[h]
    \centering\scriptsize
    \caption{These relations are utilized to describe (1) the absolute positions of objects within the layout (or container), (2) the relative positions between objects, and (3) the arrangement constraints for objects of the same category.}
    \begin{tabular}{ccc}
    \toprule
    Type & Template & Example \\
    \midrule
    Anchoring Relation & $f_{\text{anchor}}(c_i,\mathcal{B}),i\le n$ & font\_edge(plate, table)\\
    Relative Relation & $f_{\text{rel}}(\mathbf{c}_i,\mathbf{c}_j),\{i,j\}\le n$ & left\_of(fork, plate)\\
    Alignment Relation & $f_{\text{align}}(\mathbf{c_i},...,\mathbf{c_k}),\{i,k\}\le n$ & align\_x\_axis(fork, plate, knife)\\
    \bottomrule
    \end{tabular}
    \label{tab:method_relationship}
\end{table}

Due to the flexibility of natural language, the same relation can be expressed in multiple ways. Therefore, when creating new relations, it is necessary to provide the system with specific relation names to constrain its generation scope. For example, defining ``left above of" can help avoid alternatives such as ``above left of" or ``top left corner", etc.

During the generation process, the fast system further produces three components based on the specified name: the definition of the relation, the constraint function, and the validation function. The constraint function translates topological relations into numerical constraints, while the validation function evaluates whether the coordinates of object combinations align with the described topological relation.

\subsection{Coarse-grained Topological Relation Generation}

As shown in the Figure \ref{fig:method_overview} (b), after the ARL is constructed by the fast system, the language instructions, object list, and ARL are further utilized in Stage 1 to construct a set of topological relations that describe the layout. The process is detailed in the accompanying algorithm \ref{alg:stage_1}.

\begin{algorithm}\scriptsize
\caption{Stage-1 Coarse-grained Generation}
\begin{algorithmic}[1] 
\State \textbf{Input:}  $ I, O, ARL$ 
\State \textbf{Output:}  $ \mathcal{C}, \mathcal{R} $ 

\State  $ \text{SceneDesc} \leftarrow \text{RRG}(I, O) $ 
\State  $ \mathcal{C} \leftarrow \text{GenDiscreteCoord}(\text{SceneDesc}) $ 
\State  $ \mathcal{R} \leftarrow \text{ExtractRelations}(\text{SceneDesc}) $ 
\State  $ \text{validity} \leftarrow \text{Consistency}(\mathcal{C}, \mathcal{R}) $ \Comment{relations Filtering}
\State $k \leftarrow \text{Select} \ index \ \text{where} \ \text{validity is False}$ \Comment{Functionally Incomplete}
\For{$i$ \textbf{in} $k$}
    \State  $ \mathcal{C}, \mathcal{R} \leftarrow \text{Update}(\mathcal{C}, \mathcal{R}, O_i) $ 
\EndFor
\State \textbf{return}  $ \mathcal{C}, \mathcal{R} $ 
\end{algorithmic}
\label{alg:stage_1}
\end{algorithm}

First, high-quality expected scene descriptions are obtained through the three-stage processing of the slow system: Reasoning, Reflection, and Generation (RRG). Based on the preferences embedded in the language instructions and the attributes of the objects, the system identifies functional areas, determines object interactions, and generates the expected positions of each object. Subsequently, reflection is performed to ensure all objects are adequately described, potential conflicts are resolved, and a complete scene description is produced.

After obtaining the scene description, the fast system is utilized to describe the topological relations of the layout in two ways: 1) \textbf{Discrete Coordinates}: The layout is discretized into a grid, and the fast system places each object according to the description, generating a coordinate set $\mathcal{C}$. 2) \textbf{Semantic Relations}: The topological relations of the layout are described based on the definitions provided in the ARL, and regularization is applied to derive $\mathcal{R}$.
Subsequently, inconsistent relations are filtered using the validation functions in the ARL, and the positions of missing objects are redeployed using the fast system.

\subsection{Fine-grained Physical Position Grounding}

\begin{algorithm}\scriptsize
\caption{Stage-2 Fine-grained Grounding}
\begin{algorithmic}[1] 
\State \textbf{Input:}  $ \mathcal{C}, \mathcal{R}, \text{obj\_sizes } \mathcal{S}, \text{table } \mathcal{B}, {ARL} $ 
\State \textbf{Output:}  $ \text{best\_points } C_p^*, \text{best\_boxes } C_b^* $ 
\State  $ \Lambda \leftarrow \text{load\_or\_generate\_relation\_funcs}(ARL) $ 
\State  $ \text{Pop} \leftarrow \{\text{random\_feasible\_layout}(\mathcal{C})\} \times N $ 
\For{ $ t = 1 $  \textbf{to}  $ T $ }
    \State  $ \text{Fit} \leftarrow [\text{fitness}(p) \text{ for } p \text{ in Pop}] $  
    \State  $ \text{Pop} \leftarrow \text{select\_top\_}\frac{1}{2}(\text{Pop}, \text{Fit}) $ 
    \State  $ \text{Pop} \leftarrow \text{crossover\_and\_mutate}(\text{Pop}) $  \Comment{0.3 ratio of mutation}
    \If{ $ \max(\text{Fit}) == \text{full\_score} $ }
        \State \textbf{break}
    \EndIf
\EndFor
\State  $ C_p^* \leftarrow \text{normalize}(\text{argmax\_Fit}(\text{Pop})) $ 
\State  $ C_b^* \leftarrow \text{to\_bbox}(C_p^*, \mathcal{S}, \mathcal{B}) $ 
\State \textbf{return}  $ C_p^*, C_b^* $ 
\end{algorithmic}
\end{algorithm}

In the second stage, continuous optimization is performed based on the discrete layout generated in Stage 1. The desktop is mapped to a 600 $\times$ 300 pixel plane, with each object's position encoded as the coordinates of its top-left corner. A genetic algorithm with 2,000 individuals and 100 generations is employed to search for the optimal solution. The fitness function comprises two components: \textbf{Physical feasibility}: ensuring collision-free placement and stability. \textbf{Semantic consistency}: evaluated by invoking constraint functions from the ARL for each relation.
Additionally, during the generation process, if any relation function is missing or produces an error, the fast system promptly calls an LLM to generate or repair the Python code and dynamically reload it, ensuring uninterrupted optimization and continuous self-healing of the ARL.

The selection-crossover-mutation process follows a pipeline of ``retaining the top 50\% $\to$ gene-level random inheritance $\to$ two types of mutation (random reinitialization or Gaussian perturbation)." Iterations are either terminated after a fixed number or earlier if all constraints achieve maximum scores. The final output consists of a set of normalized pixel coordinates and bounding boxes.

\subsection{Self-Validation}

Throughout the entire process, three stages of self-validation were implemented. In the first stage, consistency filtering based on $\mathcal{C}$ and $\mathcal{R}$ is applied as self-validation to remove ambiguous relations. Subsequently, the validation function within the ARL is invoked to assess layout quality. Any potential incompleteness in the layout caused by the removal of relations was addressed using the fast system, constituting the second stage of self-validation.

The third and most critical stage was designed to address potential failures in dynamically generated constraint functions, which may result from improper threshold settings.
As indicated by the red flow in the Figure \ref{fig:method_overview}, after the second-stage layout optimization, validation is performed using the validation functions from the ARL. For each relation  $r \in R$, the system executes the corresponding validation function $\text{is}\_r(L)$ to determine whether the current layout $C$ satisfies $r$. If $\text{is}\_r(L)$ returns ``\textbf{False}", the validation is considered to have failed.

When validation failure is detected, the current layout $C$ and the failed relation $r$ are fed back to the large model. The model then adjusts the parameters of the constraint function associated with $r$ (e.g., the thresholds of collision recognition) based on historical inputs and feedback, and updates the dynamic relation library accordingly. The feedback-adjustment process is designed to be conservative to ensure that larger errors are not introduced. The refined layout is then re-entered into the fine-tuning phase, and this process iterates until all validation functions return positive results.

This closed-loop mechanism ensures that the final output layout meets the predefined requirements in terms of physical feasibility, semantic consistency, and aesthetic quality.
 
 \section{Experiment}
In this section, the following three questions are explored through experiments:
\textbf{Q1}: Does AutoLayout demonstrate superior performance in layout generation, including physical plausibility, semantic consistency, and functional completeness?
\textbf{Q2}: With the integration of ARL, does our method exhibit the potential to generalize to other layout generation tasks?
\textbf{Q3}: Are the proposed slow-fast system and self-vaildation mechanism critical for addressing this task effectively?

\subsection{Experimental Setup}

\textbf{Evaluation.} The models were evaluated based on the following criteria: (1) performance in physical plausibility, semantic consistency, and functional completeness of the generated layouts; (2) ability to process open-vocabulary language instructions; and (3) generalizability to novel scenarios.
Test cases were created across 8 desktop types, with each type featuring 3 asset combinations, each containing at least 5 categories. All methods utilized the same pre-processed resources. Details of test case generation are provided in the Appendix.

\textbf{Evaluation Metrics.} The evaluation metrics assess three key aspects: physical plausibility, semantic consistency, and functional completeness.
Physical plausibility includes two indicators: the \textit{Collision-Free Score (CF)} and the \textit{In-Boundary Score (IB)}. The CF measures whether objects exhibit any unreasonable collisions or overlaps, while the IB determines whether all objects remain properly within their designated containers or platforms.
Semantic consistency uses GPT-4o as the evaluator to determine whether the generated 3D layouts align with the input language instructions, including position (Pos.) and alignment (Ali.).
\textit{Functional completeness (FC)} directly evaluates whether all proposed objects in the layout are appropriately placed.
A comprehensive evaluation metric, the \textit{Physical-Semantic-Functional Score (PSF)}, combines these three aspects into a weighted aggregate score. More details can be found in the Appendix.



\textbf{Baselines.} In this paper, state-of-the-art layout generation methods, including LayoutGPT \cite{feng2023layoutgpt}, HoloDeck \cite{yang2024holodeck}, and I-Design \cite{ccelen2024design}, is used as baselines. However, as these methods is originally designed for indoor scene generation, their prompts are adapted based on the examples provided in the context to suit the requirements of our task. The LLMs used in among the approaches is GPT-4o. 

\subsection{Quantitative Analysis}

\begin{table}[]
    \centering\scriptsize
    \setlength{\tabcolsep}{2.5pt}
    \caption{Average Performance over 8 Desktop Scenarios.}
    \begin{tabular}{c|cccccc|cccccc|cccccc|}
        \toprule
         \multicolumn{1}{c}{} & \multicolumn{6}{c}{\textbf{Dinning Table}} & \multicolumn{6}{c}{\textbf{Tea break table}} & \multicolumn{6}{c}{\textbf{Office Table}} \\
        \cmidrule(lr){2-7} \cmidrule(lr){8-13}\cmidrule(lr){14-19}
        Methods & CF & IB & Pos. & Ali. & FC & \cellcolor{red!10} PSF & CF & IB & Pos. & Ali. & FC & \cellcolor{red!10} PSF& CF & IB & Pos. & Ali. & FC & \cellcolor{red!10} PSF \\
        \midrule
        LayoutGPT   & 71.4 & 75.8 & \textbf{75.0} & 55.3 & \textbf{100.0} & \cellcolor{red!10} 79.0 & 77.4 & 61.5 & 76.0 & 67.0 & 84.6 & \cellcolor{red!10} 74.6 & 63.5 & 59.3 & 75.7 & 78.7 & 86.7 & \cellcolor{red!10} 73.7\\ 
        HOLODECK    & 85.6 & 87.5 & 70.0 & 59.0 & 88.9 & \cellcolor{red!10} 80.6 & 95.2 & 91.7 & 65.3 & 51.0 & 50.4 &\cellcolor{red!10} 69.9 & 77.8 & 80.0 & 58.3 & 70.0 & 54.4 &\cellcolor{red!10} 67.1\\ 
        I-Design    & 67.3 & 70.8 & 74.3 & 63.0 & 83.3 & \cellcolor{red!10} 73.2 & 75.4 & 71.3 & 73.7 & 54.3 & 61.4 &\cellcolor{red!10} 67.0 & 77.8 & 73.9 & 69.3 & 68.3 & 58.1 & \cellcolor{red!10} 68.4\\ 
        AutoLayout  & \textbf{100.0} & \textbf{100.0} & \textbf{75.0} & \textbf{73.7} & \textbf{100.0} &\cellcolor{red!10} \textbf{92.3} & \textbf{100.0} & \textbf{100.0} & \textbf{81.7} & \textbf{88.0} & \textbf{100.0} &\cellcolor{red!10} \textbf{95.5} & \textbf{95.2} & \textbf{94.4} & \textbf{81.3} & \textbf{86.0} & \textbf{100.0} &\cellcolor{red!10} \textbf{93.0}\\ 
       \hline
       \hline
        \multicolumn{1}{c}{} & \multicolumn{6}{c}{\textbf{Study Table}} & \multicolumn{6}{c}{\textbf{Dressing Table}} & \multicolumn{6}{c}{\textbf{Craft Table}} \\
        \cmidrule(lr){2-7} \cmidrule(lr){8-13}\cmidrule(lr){14-19}
        Methods & CF & IB & Pos. & Ali. & FC & \cellcolor{red!10}PSF & CF & IB & Pos. & Ali. & FC & \cellcolor{red!10}PSF& CF & IB & Pos. & Ali. & FC & \cellcolor{red!10}PSF \\
        \midrule
        LayoutGPT   & 85.9 & 63.9 & 76.7 & 78.0 & 100.0 &\cellcolor{red!10} 80.9 & 77.8 & 76.7 & \textbf{82.7} & \textbf{70.0} & \textbf{100.0} &\cellcolor{red!10} 83.8 & 83.8 & 58.9 & \textbf{78.3} & 61.3 & \textbf{100.0} &\cellcolor{red!10} 79.5\\ 
        HOLODECK    & 75.6 & 72.4 & 67.7 & 80.7 & 51.5 &\cellcolor{red!10} 67.3 & 71.1 & 83.8 & 76.0 & 57.3 & 80.0 &\cellcolor{red!10} 75.0 & 70.0 & 75.9 & 66.7 & 57.0 & 83.3 &\cellcolor{red!10} 72.7\\ 
        I-Design    & 72.2 & 73.9 & 69.3 & 68.3 & 58.1 &\cellcolor{red!10} 67.3 & 86.7 & 93.3 & 63.3 & 53.7 & 80.0 &\cellcolor{red!10} 77.6 & 72.5 & 71.9 & 75.0 & 47.3 & 93.3 &\cellcolor{red!10} 75.2\\ 
        AutoLayout  & \textbf{100.0} & \textbf{95.2} & \textbf{81.0} & \textbf{85.3} & \textbf{100.0} &\cellcolor{red!10} \textbf{94.0} & \textbf{95.6} & \textbf{100.0} & 61.7 & 66.0 & \textbf{100.0} &\cellcolor{red!10} \textbf{88.3} & \textbf{100.0} & \textbf{100.0} & 75.0 & \textbf{74.7} & \textbf{100.0} & \cellcolor{red!10}\textbf{92.5} \\ 
       \hline
       \hline
        \multicolumn{1}{c}{} & \multicolumn{6}{c}{\textbf{Fruits Table}} & \multicolumn{6}{c}{\textbf{Bar Table}} & \multicolumn{6}{c}{\textbf{Average}} \\
        \cmidrule(lr){2-7} \cmidrule(lr){8-13}\cmidrule(lr){14-19}
        Methods & CF & IB & Pos. & Ali. & FC &\cellcolor{red!10} PSF & CF & IB & Pos. & Ali. & FC & \cellcolor{red!10}PSF& CF & IB & Pos. & Ali. & FC &\cellcolor{red!10} PSF \\
        \midrule
        LayoutGPT   & 88.0 & 88.9 & 67.3 & 51.0 & 90.5 &\cellcolor{red!10} 80.3 & 92.5 & \textbf{100.0} & \textbf{79.3} & 48.0 & 68.1 &\cellcolor{red!10} 78.0 & 80.0 & 73.1 & \textbf{76.4} & 63.7 & 90.3 &\cellcolor{red!10} 78.7\\ 
        HOLODECK    & 88.9 & 87.8 & 58.3 & 45.0 & 59.3 & \cellcolor{red!10}68.6 & 80.0 & 82.2 & 64.0 & 40.3 & 44.5 & \cellcolor{red!10}61.4 & 80.5 & 82.7 & 65.8 & 57.5 & 64.0 &\cellcolor{red!10} 70.3\\ 
        I-Design    & 77.8 & 80.0 & 65.0 & 45.3 & 58.7 &\cellcolor{red!10} 65.7 & 59.8 & 66.7 & 62.3 & 47.3 & 55.2 &\cellcolor{red!10} 58.3 & 73.7 & 74.9 & 69.0 & 55.7 & 68.9 &\cellcolor{red!10} 69.1\\ 
        AutoLayout  & \textbf{99.1} & \textbf{100.0} & \textbf{70.7} & \textbf{57.7} & \textbf{100.0} &\cellcolor{red!10} \textbf{89.1} & \textbf{100.0} & \textbf{100.0} & 68.3 & \textbf{57.0} & \textbf{100.0} &\cellcolor{red!10} \textbf{88.8} & \textbf{98.7} & \textbf{98.7} & 74.3 & \textbf{73.5} & \textbf{100.0} &\cellcolor{red!10} \textbf{91.7}\\ 
    \bottomrule
    \end{tabular}
    \label{tab:main result}
\end{table}

\textbf{Physical Plausibility}. AutoLayout attains near-perfect CF (98.7 \%) and IB (98.7 \%), reducing object overlaps and boundary violations almost entirely. By contrast, LayoutGPT averages 80.0 \% CF and 73.1 \% IB—reflecting frequent clumping or off-table placements—while HOLODECK (80.5 \% CF, 82.7 \% IB) and I-Design (73.7 \% CF, 74.9 \% IB) also suffer from residual collisions and out-of-bounds errors. The closed-loop validation mechanism in AutoLayout operates across multiple stages to detect and resolve physical conflicts. It resamples topological relations and discrete coordinates, adjusts constraint thresholds within the ARL, and refines object poses using an evolutionary algorithm. This process ensures that nearly all generated layouts are collision-free and fully inclusive.

It is worth noting that, compared to single-stage approaches such as LayoutGPT and programmatic sketch design, the two-stage holographic deck demonstrates superior performance in terms of physical plausibility. However, it struggles to handle collisions and boundary constraints when placing a large number of complex-shaped objects in constrained spaces, such as fruits table and bar table. In contrast, our closed-loop framework effectively fine-tunes these challenging scenarios, providing a more robust solution.

\textbf{Semantic Consistency}.
In terms of Pos. and Ali. metric, AutoLayout achieved 74.3\% in Pos. and 73.5\% in Ali., outperforming HOLODECK (65.8\% Pos., 57.5\% Ali.) and I-Design (69.0\% Pos., 55.7\% Ali.). However, AutoLayout showed slightly lower performance in Pos. compared to LayoutGPT, likely due to LayoutGPT benefiting from aesthetic guidance provided by contextual examples. Nevertheless, due to the lack of physical constraints in LayoutGPT, its performance in alignment metrics was inferior to AutoLayout.
Leveraging the RRG pipeline in the slow system, AutoLayout explicitly performs relational reasoning through the ARL before being grounded by the fast system. This approach demonstrates more robust performance compared to fixed-rule methods, reducing the rate of misaligned or misplaced objects by over 10\%.

\textbf{Functional Completeness}.
AutoLayout places every requested object in every scenario (FC = 100 \%), whereas LayoutGPT omits objects in 9.7 \% of cases on average, and HOLODECK and I-Design drop 36 \% and 31 \% of items respectively under complex or contradictory constraints. The integrated self-validation loop explicitly checks for missing objects after each filtering or optimization pass, triggering re-injection and relocation until full coverage is achieved without manual intervention.

Aggregating across all metrics, AutoLayout achieves a PSF score of 91.7 \%, a 13.0–22.6 point improvement over the best baseline (LayoutGPT, 78.7 \%; HOLODECK, 70.3 \%; I-Design, 69.1 \%). This consistent advantage across diverse tasks—from densely packed craft tables to narrow office desks—underscores the robustness of our closed-loop, slow-fast design and the generality afforded by an adaptive, LLM-driven relation library.

\subsection{Qualitative Analysis}


\subsubsection{Human Evaluation}

\begin{table}[]
    \centering\scriptsize
    \caption{Performance on Variable Complexity Levels (Dining and Office Tables).}
    \begin{tabular}{ccccccccccc}
       \toprule
        & \multicolumn{5}{c}{Dinning Table} & \multicolumn{5}{c}{Office Table} \\
       \cmidrule(lr){2-6}\cmidrule(lr){7-11}
       Methods & Easy & Medium & Hard & Expert & Mean & Easy & Medium & Hard & Expert & Mean \\
       \midrule
       LayoutGPT   & 0 & 0 & 0 & 0 & 0 & 0 & 0 & 0 & 0 & 0 \\ 
       HOLODECK    & 0 & 0 & 0 & 0 & 0 & 0 & 0 & 0 & 0 & 0 \\ 
       I-Design    & 0 & 0 & 0 & 0 & 0 & 0 & 0 & 0 & 0 & 0 \\ 
       AutoLayout  & 1.00 & 0.929 & 0.944 & 1.00 & 0.95 & 1.00 & 0.786 & 0.944 & 0.500 & 0.875 \\ 
       \bottomrule
    \end{tabular}
    \vspace{-15pt}
    \label{tab:level_result}
\end{table}

To further evaluate the proposed method's performance across varying complexity scenarios, two of the most common real-life settings were selected: dining tables and office desks. The complexity was categorized into four levels based on the number of items on the table: simple (1–3 items), moderate (4–5 items), difficult (6–8 items), and expert (9 items). A total of 40 layouts were constructed for these two scenarios, with each layout being generated twice.

As shown in Table~\ref{tab:level_result}, the performance of each method in the two scenarios was evaluated based on assessments conducted by human experts. It shows that LayoutGPT, HOLODECK, and I-Design all scored zero, despite sometimes satisfying minimal collision/boundary requirements.  Designers cited cluttered centripetal placements and “unnatural” floating or partial overlapping as visually jarring.
In contrast, AutoLayout achieved a mean human score of 0.95 on Dining and 0.875 on Office, consistently praised for “balanced compositions” and realistic stacking.  Even under Expert-level (9 items) instructions, the method placed every object in an aesthetically coherent manner.

\begin{figure}[h]
    \centering
    \includegraphics[width=1\linewidth]{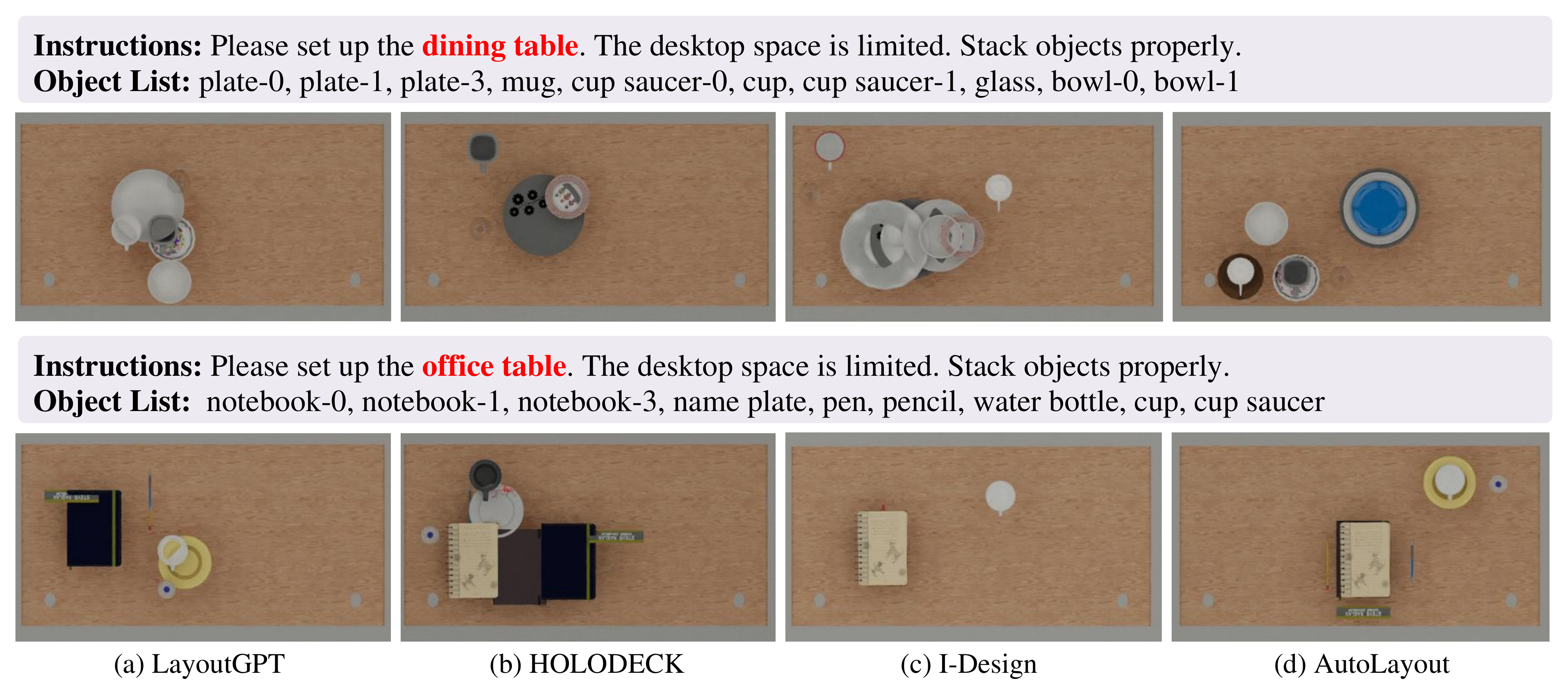}
    \vspace{-10pt}
    \caption{Visualizations of two examples with specified requirements requiring stacking}
    \vspace{-10pt}
    \label{fig:case_study}
\end{figure}

\subsubsection{Case Study}

As shown in Figure \ref{fig:case_study}, in the dining-table scenario, every baseline collapses once the instruction demands coordinated stacking. LayoutGPT ignores the ``stack objects properly” clause and simply piles the entire inventory into a single, off-center heap, triggering both collision and in-boundary violations. While Holodeck significantly reduces collision and in-boundary violations, it often bypasses the placement of many objects in challenging scenarios to simplify layout generation. This approach compromises the functional completeness of the resulting layout. Similar to LayoutGPT, I-Design attempts to stack a large number of plates and bowls without ignoring any objects. However, this often results in physically implausible configurations. In contrast, AutoLayout first decomposes the request into an RRG scene description—“central serving zone” for plates, ``vertical stack” for bowls, ``peripheral drinkware ring”—then lets the fast system instantiate these relations on a discrete grid and iteratively prunes collisions through the ARL’s learnt ``\textit{on\_top\_of}” and ``\textit{align\_x\_axis}” functions. The closed-loop validation subsequently tightens thresholds until every bowl rests stably on a plate and every cup-saucer pair nests without overlap, yielding 100 \% CF, IB, and FC scores reported in Table \ref{tab:main result}. The qualitative result therefore exhibits the rare combination of tidy composition and strict physical realism that the baselines never approached.

Similarly, the office-table case stresses semantic fidelity rather than sheer packing density: notebooks must remain readable, pens reachable, and a nameplate centred. The baseline agents misinterpreted the soft constraints. LayoutGPT placed a cup in the writing area and failed to stack the nameplate with the cup. HOLODECK ignored the stacking requirement, scattering notebooks across the surface. I-Design, on the other hand, eliminated more than half of the objects altogether. AutoLayout’s slow system resolves such functional semantics explicitly: it anchors the nameplate to the upper midline, aligns three notebooks along the z-axis by ARL-generated ``\textit{align\_z-axis\_at\_center}” constraints, and reserves radial clearance around writing utensils through ``\textit{near\_of}” exclusion zones. The evolutionary grounding stage then fine-tunes positions in continuous space, while self-validation re-injects any displaced pen until the full constraint set is satisfied. The final layout secures the highest human-study score (0.875) in Table \ref{tab:level_result}, illustrating how the slow-fast collaboration not only averts spatial hallucination but also internalises nuanced affordances that previous two-stage methods could not encode.

\subsection{Ablation}

\begin{table}[h]
    \centering\scriptsize
    \caption{Ablation on Key Components. Additionally, ``Round Num" refers to the average number of sampling rounds during the generation process. If a sample cannot be solved after 5 rounds of sampling during the generation process, it is considered to have no solution.}
    \begin{tabular}{cccccccc}
        \toprule
        & \multicolumn{2}{c}{Physics} & \multicolumn{2}{c}{Semantics} & Functional & Overall Score & Round Num\\
       \cmidrule(lr){2-3}\cmidrule(lr){4-5}\cmidrule(lr){6-6}\cmidrule(lr){7-7}\cmidrule(lr){8-8}
       Methods & CF & IB & Pos. & Ali. & FC & PSF \\
       \midrule
       AutoLayout  & 99.7 & 99.4 & 78.0 & 75.3 & 100.0 & 92.8 & 1.04 \\ 
       w/o ARL & 93.2 & 97.5 & 76.9 & 75.0 & 100.0 & 90.9 & 1.38\\ 
       w/o Description  & 90.8 & 97.3 & 76.5 & 70.9 & 100.0 & 89.7 & 1.79\\ 
       w/o self-validation & 81.3 & 83.3 & 61.5 & 59.8 & 83.5 & 81.1 & 2.38\\
       w/o $\mathcal{C}$  & 88.1 & 89.7 & 70.7 & 70.0 & 100.0 & 86.7 & 1.55\\ 
       \bottomrule
    \end{tabular}
    \label{tab:ablation_component}
\end{table}


The ablation study presented in Table \ref{tab:ablation_component} demonstrates that each component of the AutoLayout framework is indispensable. Removing the ARL (w/o ARL) resulted in a 6\% drop in collision-free accuracy and increased the average number of sampling iterations from 1.04 to 1.38. This highlights that handwritten rules cannot match the real-time, code-level corrections provided by ARL. Eliminating the scene description module (w/o description) further pushed the search process into trial-and-error, causing CF metric to drop below 91\% and nearly doubling the iterations required, as the optimizer struggled to reconcile mismatched discrete coordinates with topological relations. When the closed-loop self-validation is disabled, severe failures occurred: physical plausibility, semantic consistency, and functional completeness all experienced significant declines (PSF drop $\approx$ 12\%), and the solver required more than twice the number of iterations. Finally, skipping the discrete initialization step (w/o $\mathcal{C}$ left the system unable to detect obvious geometric contradictions, leading to approximately 11\% losses in both CF and IB metrics.

In summary, these trends underscore the proposed design hypothesis: robust layout synthesis can only emerge when slow, deliberate reasoning is seamlessly integrated with fast, self-validation execution. The ARL compresses linguistic relations into deterministic spatial relations, discrete coordinates prune infeasible topological relations, and the self-validation closed loop iteratively adjusts constraints until the layout satisfies physical, semantic, and functional requirements.
Removing any of these modules disrupts this synergy, forcing the optimizer to pursue infeasible solutions and significantly increasing computation time. Therefore, the complete AutoLayout stack is not merely a collection of optional plugins but a tightly coupled system designed to maximize both accuracy and efficiency.




 \section{Conclusion}
The proposed AutoLayout framework integrates slow reasoning with fast optimization, closed-loop self-validation, and an adaptive relation library. This approach effectively achieves a unified breakthrough in physical feasibility, semantic consistency, and functional completeness, offering an efficient and robust solution for automatic layout synthesis tasks.
 
\appendix

\section{Experimental Setup}

\subsection{Test Case}

\begin{tcolorbox}[
    colback=yellow!10,      
    colframe=yellow!10,       
    width=\textwidth,     
    boxrule=0.5pt,        
    arc=1mm,              
    left=2mm, right=2mm,  
    top=2mm, bottom=2mm   
]
\small{
\begin{verbatim}
{
    "Dining_Table": {
        "scene": "Dining Table",
        "info": "Arrange a scene suitable for family or banquet dining",
        "case": [
            ["plate", "glass", "knife", "fork", "spoon", "napkin"],
            ["plate", "bowl", "chopstick", "spoon", "cup", "napkin"],
            ["plate", "glass", "knife", "fork", "candleholder", "decoration"]
        ]
    },
    "Tea_Break_Table": {
        "scene": "Tea Break Table",
        "info": "Casual scene for afternoon tea and snacks",
        "case": [
            ["plate", "snack box", "glass", "cake", "fork", "napkin"],
            ["mug", "cup saucer", "spoon", "snack box", "tissue", "cake", 
                "plate"],
            ["cup-0", "cup saucer-0", "cup-1", "cup saucer-1", "tea pot", 
                "snack bag-0", "snack bag-1"],
        ]
    },
    "Office_Desk": {
        "scene": "Office Desk",
        "info": "Desktop arrangement for daily office environment",
        "case": [
            ["monitor", "notebook", "pen", "tape", "paper", "cup"],
            ["notebook", "folder", "marker", "pen", "RC", "tissue"],
            ["paper-0", "paper-1", "notebook-0", "notebook-1", "coffee cup", 
                "tape"]
        ]
    },
    "Study_Desk": {
        "scene": "Study Desk",
        "info": "Study and reading scene for students",
        "case": [
            ["notebook", "pencil", "pen", "paper", "cup", "lamp"],
            ["notebook", "lamp", "mug", "tissue", "plant"],
            ["notebook", "pencil-0", "pencil-1", "model", "decoration", "lamp", 
                "plant"]
        ]
    },
    "Dressing_Table": {
        "scene": "Makeup Table",
        "info": "Desktop arrangement for makeup and skincare",
        "case": [
            ["mirror", "lamp", "tissue", "cosmetic", "decoration"],
            ["candle", "candleholder", "vase", "plant", "tissue"],
            ["mirror", "cosmetic", "tissue", "cosmetic box", "decoration"]
        ]
    },
    "Craft_Table": {
        "scene": "Craft Table",
        "info": "Workbench for handicraft production",
        "case": [
            ["paper", "scissors", "tape", "notebook", "pen"],
            ["model", "tape", "scissors", "tray", "box", "tissue"],
            ["notebook", "pen", "scissors", "tray", "tissue", "decoration"]
        ]
    },
\end{verbatim}
}
\end{tcolorbox}
\begin{tcolorbox}[
    colback=yellow!10,      
    colframe=yellow!10,       
    width=\textwidth,     
    boxrule=0.5pt,        
    arc=1mm,              
    left=2mm, right=2mm,  
    top=2mm, bottom=2mm   
]
\small{
\begin{verbatim}
    "Fruits_Table": {
        "scene": "Fruits Table",
        "info": "Table for Fruits",
        "case": [
            ["apple-0", "apple-1", "vase", "pear", "banana"],
            ["banana", "pear", "cup", "decoration", "plant", "cup saucer"],
            ["candle", "candleholder", "decoration", "plant", "lemon", "apple", 
                "pear", "peach"]
        ]
    },
    "Bar_Table": {
        "scene": "Bar Table",
        "info": "Gather wine bottles, drinking utensils and snacks to create 
            a warm private bar atmosphere",
        "case": [
            ["wine", "glass-0", "snack bag", "napkin", "glass-1"],
            ["wine-0", "wine-1", "glass", "snack box-0", "snack box-1", "tissue"],
            ["wine-0", "wine-1", "wine-2", "glass", "snack bag", "snack box", 
                "decoration", "napkin"]
        ],
    }
}
\end{verbatim}
}
\end{tcolorbox}

The above are the test cases used in the main experiment in the text, which include various scenarios and their corresponding object combinations.

\subsection{Metric}

\subsubsection{Collision-Free Score (CF)}
Similar to previous work, we employ a simple, yet effective metric to quantify pairwise collisions among $N$ rigid objects in 3D, based on the Intersection-over-Union (IoU) of their axis-aligned bounding boxes.

\begin{itemize}
  \item Let  
  \begin{equation}
    O = \{o_1, \dots, o_N\}
  \end{equation}  
  be the set of objects.  
  \item Each object $o_i$ is enclosed by an axis-aligned bounding box  
  \begin{equation}
    C_{bi} = [\min_i,\max_i], 
    \quad 
    \min_i = p_i - \frac{1}{2}s_i,\;\;
    \max_i = p_i + \frac{1}{2}s_i,
  \end{equation}  
  where  
  \begin{itemize}
    \item $p_i\in\mathbb{R}^3$ is the box center,  
    \item $s_i\in\mathbb{R}^3$ is its size vector.
  \end{itemize}
\end{itemize}

For each unordered pair $(i,j)$, define:

\begin{enumerate}
  \item \textbf{Intersection box}  
  \begin{equation}
    I_{ij} 
    = C_{bi} \cap C_{bj}
    = \bigl[\max(\min_i,\min_j),\min(\max_i,\max_j)\bigr].
  \end{equation}
  
  \item \textbf{Intersection volume}  
  \begin{equation}
    V(I_{ij})
    = \prod_{k=1}^3 \max\{0,(I_{ij})_{\max}^k - (I_{ij})_{\min}^k\}.
  \end{equation}
  
  \item \textbf{Union volume}  
  \begin{equation}
    V(U_{ij})
    = V(C_{bi}) + V(C_{bj}) - V(I_{ij}), 
    \quad
    V(C_{bi})=\prod_{k=1}^3 s_i^k.
  \end{equation}
  
  \item \textbf{IoU}  
  \begin{equation}
    \mathrm{IoU}_{ij}
    = \frac{V(I_{ij})}{V(U_{ij})}
    \in[0,1].
  \end{equation}
\end{enumerate}

\begin{itemize}
  \item Let $\tau\geq 0$ be an IoU threshold.  
  \item Define the set of \textbf{colliding pairs}  
  \begin{equation}
    \mathcal{C}(\tau) 
    = \{(i,j)\mid i<j,\;\mathrm{IoU}_{ij}>\tau\},
    \quad C = |\mathcal{C}(\tau)|.
  \end{equation}
  \item Total number of pairs  
  $\displaystyle M = \frac{1}{2}N(N-1)$.
\end{itemize}

\begin{enumerate}
  \item \textbf{Mean IoU over collisions}  
  \begin{equation}
    \overline{\mathrm{IoU}}
    = 
    \begin{cases}
      \frac{1}{C}\sum_{(i,j)\in\mathcal{C}(\tau)}\mathrm{IoU}_{ij}, 
        & C>0,\\
      0, & C=0.
    \end{cases}
  \end{equation}
  
  \item \textbf{Collision-Free Score (CF)}  
  \begin{equation}
    \mathrm{CF}
    = 1 - \frac{C}{M}
    \in[0,1],
    \quad
    \mathrm{CF}=1\;\Leftrightarrow\;C=0.
  \end{equation}
  
  \item \textbf{Positive Ratio (Collision Rate)}  
  \begin{equation}
    \rho = \frac{C}{M}.
  \end{equation}
\end{enumerate}

Notably, we set the $\tau$ as 0.01

\subsubsection{In-Boundary Score (IB)}

Similarly, we also calculated whether these objects are within the appropriate boundary range and measured this constraint.
Let $N$ be the number of objects (excluding the table).  We assume each object $o_i$ has axis-aligned bounding box $\mathcal{C}_{bi}$, and the supporting table has box $\mathcal{B}$.

For each $i$, define

\begin{equation}
v_i^T =
\begin{cases}
1, & \mathcal{C}_{bi}\not\subset\mathcal{B}\text{ in }x,y
\quad\text{or}\quad
\min\nolimits_{z}\bigl(\mathcal{C}_{bi}\bigr)<\min\nolimits_{z}(\mathcal{B}),\\
0, & \text{otherwise.}
\end{cases}
\end{equation}

The total table-containment violations:

\begin{equation}
V_T \;=\;\sum_{i=1}^N v_i^T.
\end{equation}

An unordered pair $(i,j)$ is a \textbf{stacking pair} if
\begin{equation}
\mathrm{proj}_{xy}(\mathcal{C}_{bi})\;\cap\;\mathrm{proj}_{xy}(\mathcal{C}_{bj})
\neq \varnothing
\quad\text{and}\quad
\bigl(\max\nolimits_{z}(\mathcal{C}_{bi})\le\min\nolimits_{z}(\mathcal{C}_{bj})
\;\vee\;
\max\nolimits_{z}(\mathcal{C}_{bj})\le\min\nolimits_{z}(\mathcal{C}_{bi})\bigr).
\end{equation}

Let
\begin{equation}
S \;=\;\bigl|\{(i,j):i<j\text{ is a stacking pair}\}\bigr|.
\end{equation}

For each stacking pair, denote by $b$ the bottom and $t$ the top object, and set

\begin{equation}
v^{S}_{ij} =
\begin{cases}
1, & \mathrm{proj}_{xy}(\mathcal{C}_{b_b})\not\supset
      \mathrm{proj}_{xy}(\mathcal{C}_{b_t}),\\
0, & \text{otherwise.}
\end{cases}
\end{equation}

The total incomplete-containment violations:
\begin{equation}
V_S \;=\;\sum_{\substack{(i,j)\\\text{stacking}}} v^S_{ij}.
\end{equation}

Then, we can calculate the IB score as follow:

\begin{equation}
    \mathrm{IB} = \frac{V_T+V_S}{N + S}
\end{equation}

\subsubsection{Position (Pos.) \& Alignment (Ali.)}

Similar to AutoVLM,we employ GPT4o as an evaluator to score the generated layouts. We input the rendered images along with the task instructions and assess whether each object is placed in an appropriate position (Pos. score). In addition,we also evaluate whether the overall layout of the objects in the layout is sufficiently neat, regardless of whether their positions are appropriate (Ali. score).

\subsubsection{Functional completeness (FC)}

We assess how completely a scene layout satisfies the object, which presence requirements of a natural language instruction.

\begin{equation}
  \mathrm{FC}
  = \frac{\hat{N}}{N}
\end{equation}

where $N$ is the number of the objects mentioned in the instruction, and $\hat{N}$ is the number of the objects generated in the layout.
 
\subsubsection{Physical-Semantic-Functional Score (PSF)}

Based on the metrics calculated above, we have obtained a series of scores that evaluate the quality of the layout from various dimensions. Therefore, we combine these scores in the following manner to obtain a comprehensive evaluation metric.

\begin{equation}
PSF = 0.4 * (\frac{CF+IB}{2}) + 0.3 * (\frac{Pos. + Ali.}{2}) + 0.3 * FC
\end{equation}

\section{More Experiment}

\subsection{Ablation of API}
To further validate the effectiveness of the slow-fast system,we conducted ablation experiments on the LLMs that constitute the slow-fast system.In the slow system, we directly employed the state-of-the-art reasoning LLM to generate scene descriptions, instead of relying on a general LLM to generate them within the RRG pipeline. As shown in Table \ref{tab:ablation_api}, we utilized GPT-o4-mini and DeepSeek-R1 to generate the dining table and office desk scenarios in the main text, respectively. We can observe that the performance difference between them is not significant, indicating that our method, empowered by RRG, can match the performance of more powerful reasoning models. In particular, we found that the use of reasoning models significantly increases the generation time. Additionally, we can also observe that our method effectively reduces the probability of retries (lower round number), thereby further improving the efficiency of data generation.

Similarly, we replaced GPT4o in the fast system with Grok-3 and Qwen-Plus. The results are shown in the table, with noticeable differences in performance. In particular, Qwen-Plus experienced a significant performance loss. Although Grok had a minimal loss, the number of retries increased significantly. This also reflects that the fast system relies more on the spatial reasoning capabilities of the LLM itself.

\begin{table}[h]
    \centering\scriptsize
    \caption{Ablation of slow-fast system}
    \begin{tabular}{ccccccccc}
       \toprule
       \multicolumn{2}{c}{AutoLayout} & \multicolumn{2}{c}{Physics} & \multicolumn{2}{c}{Semantics} & Functional & Overall Score & Round Num\\
       \cmidrule(lr){1-2}\cmidrule(lr){3-4}\cmidrule(lr){5-6}\cmidrule(lr){7-7}\cmidrule(lr){8-8}\cmidrule(lr){9-9}
       Slow & Fast & CF & IB & Pos. & Ali. & FC & PSF \\
       \midrule
       GPT4o & GPT4o & 99.7 & 99.4 & 78.0 & 75.3 & 100.0 & 92.8 & 1.04\\
       \midrule
       \multicolumn{8}{l}{\textit{Ablating the slow system}} \\
       \midrule
       GPT-o4-mini & GPT4o & 98.5 & 97.5 & 76.3 & 76.3 & 98.8 & 91.7 & 1.28\\
       DeepSeek-R1 & GPT4o & 99.8 & 99.3 & 77.6 & 77.2 & 100.0 & 93.0 & 1.21 \\
       \midrule
       \multicolumn{8}{l}{\textit{Ablating the fast system}} \\
       \midrule
       GPT4o & Grok-3 & 98.9 & 98.8 & 76.5 & 75.9 & 100.0 & 92.4 & 1.88   \\
       GPT4o & Qwen-Plus & 94.9 & 94.1 & 70.4 & 69.1 & 95.0 & 87.2 & 1.48 \\
       \bottomrule
    \end{tabular}
    \label{tab:ablation_api}
\end{table}


\subsection{More Visualizations}

We present more data samples generated in the Main Result section. It can be clearly observed that our method is capable of arranging the required objects neatly and reasonably. In contrast,other methods may encounter issues of physical implausibility and semantic inconsistency. More critically, in some samples, objects are omitted to reduce the complexity of arrangement.

\begin{figure}[h]
    \centering
    \includegraphics[width=1\linewidth]{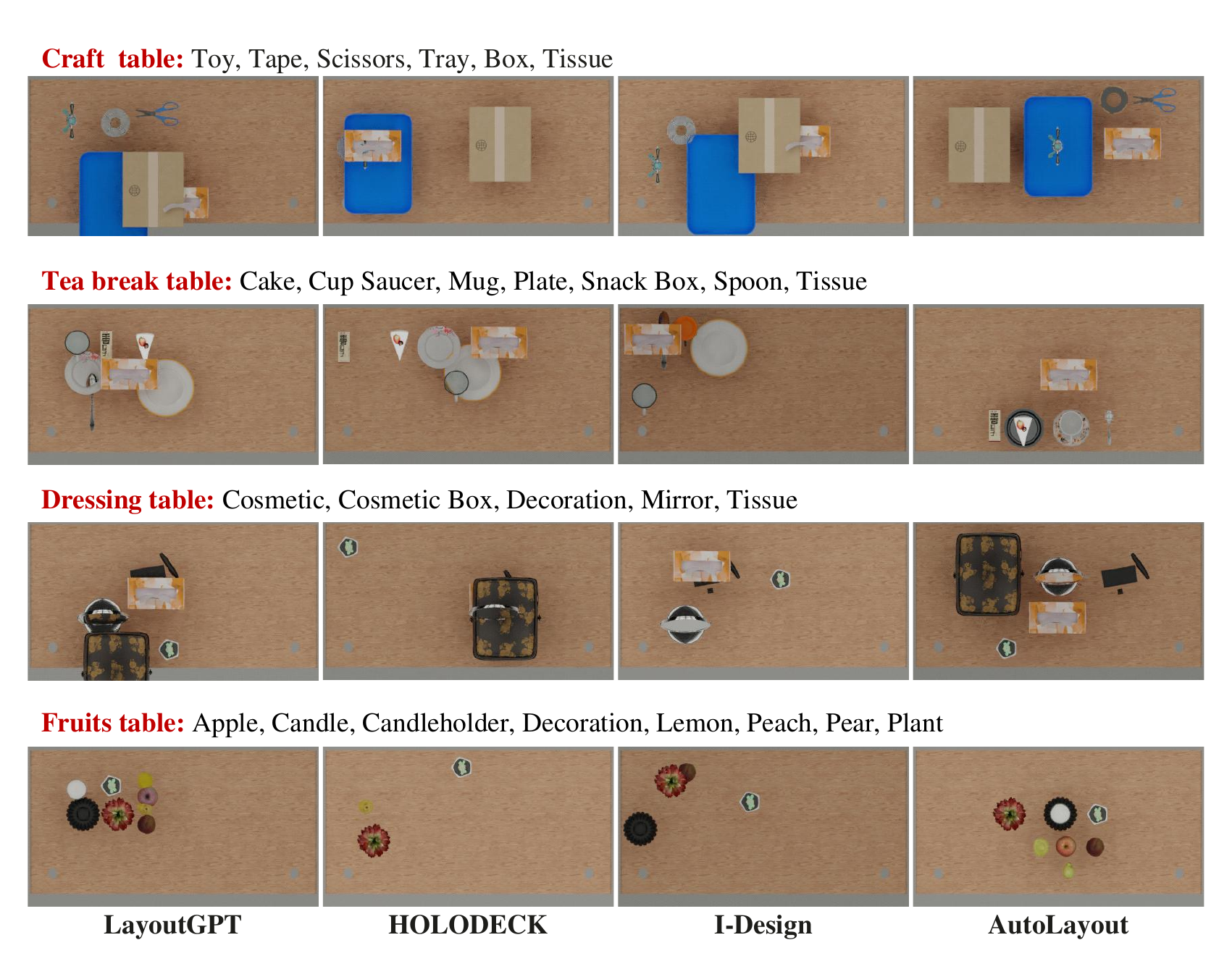}
    \caption{Visualization of the four desktop setup layouts}
    \label{fig:my_label}
\end{figure}

\section{Details of AutoLayout}
In this section,we will further elaborate on the details of AutoLayout to enhance the reproducibility of the method.

\subsection{Symbols}

As shown in the Table \ref{tab:app_symbols}, we provide the definitions of all the symbols mentioned in the main text.

\begin{table}[h]
    \centering
    \caption{Explanation of Symbols}
    \begin{tabular}{cl}
    \toprule
    Symbols & Definition \\
    \midrule
    $\mathcal{I}$ & A language instruction I (e.g., "Set up the table")\\
    $\mathcal{O}$ & A set of objects $\mathcal{O} = \{O_1, . . . ,O_n\}$\\
    $s_i$ & The size vector of the $i^{th}$ object \\
    $\mathcal{B}$ & The boundary of the supporting surface (e.g., a tabletop or room floor)\\
    $\mathcal{R}$ & A set of topological relation \\
    $\mathcal{C}$ & A set of objects' coordinates in the layout \\
    $\mathcal{C}_p$ & A set of normalized objects' pixel coordinates in the layout \\
    $\mathcal{C}_b$ & A set of normalized objects' bounding boxes in the layout \\
    \bottomrule
    \end{tabular}
    \label{tab:app_symbols}
\end{table}

\subsection{Adaptive Relation Library}
Before generating the layout, we first construct the Adaptive Relation Library (ARL), which involves three steps corresponding to the generation of the three parts in ARL. The first step is the \textbf{definition of relations}, where we can refer to some existing simple relations to generate more complex ones. 

\begin{tcolorbox}[
    colback=gray!10,      
    colframe=gray!10,       
    width=\textwidth,     
    boxrule=0.5pt,        
    arc=1mm,              
    left=2mm, right=2mm,  
    top=2mm, bottom=2mm   
]
You are a spatially imaginative and logical layout generation assistant.
I want to create an arrangement of objects that is both ordered (physically feasible and visually appealing) and functional (based on instructions to meet the user's intended purpose), called Active Relationships.
\\\\
However, I currently have a relationship that is incomplete and I need your help to perfect it base on it's name. I will provide some complete relationships for reference.
\\\\
\#\# Abstract Relationships\\
We provide the following abstract relational libraries, which are usually used to describe relative location relationships between objects:
<complete\_relationship>
\\\\
Notely, In a relative relationship, you need to define the relative position coordinates between objects, in the format $RPC(Obj_B, Obj_A) = (x, y, z)$
\\\\
\#\# Guidance
Distinguish whether it's above the z axis or above the y axis, which you can determine by RPC in the relationship definition.
Strictly follow the format output in the example.
\\\\
\#\# Example\\
</new\_relationship>{"right\_above\_of": {"type": "Binary", "definition": "Obj\_A is immediately to the right of and above of Obj\_B, viewed from the canonical camera frame. The left below side of Obj\_A and the right top side of Obj\_B are within a threshold, with substantial x-y-overlap", "RPC":[1, 1, 0]}
}</new\_relationship>
\\\\
\#\# You Task\\
incomplete relationship: <incomplete\_relationship>
\end{tcolorbox}

Among them, <complete\_relation> refers to some simple relations that have already been manually constructed,such as ``left\_of”. In addition, after defining the relations, we also need corresponding constraint functions. These are used to quantify the discrepancy between the current layout and the required relations, and then hand it over to the optimizer for optimization. Below is the prompt for generating these constraint functions.

\begin{tcolorbox}[
    colback=gray!10,      
    colframe=gray!10,       
    width=\textwidth,     
    boxrule=0.5pt,        
    arc=1mm,              
    left=2mm, right=2mm,  
    top=2mm, bottom=2mm,   
]
You're a thoughtful programmer, and you're good at using spatial imagination and reasoning to formulate a function that represents the constraints that an object needs to obey in a layout.\\\\
Here is a spatial relationship functions of "left\_of":\\\\
</func>\\
\small{
\begin{verbatim}
def left_of(objs, table_width=600, table_height=300):
    # Determine whether object A is on the left of object B
    obj_a, obj_b = objs
    A_min_x = obj_a[0]  # The minimum x value of A
    A_max_x = obj_a[2]  # The maximum x value of A
    B_min_x = obj_b[0]  # The minimum x value of B
    B_max_x = obj_b[2]  # The maximum x value of B
    
    A_min_y = obj_a[1]  # The minimum y value of A
    A_max_y = obj_a[3]  # The maximum y value of A
    B_min_y = obj_b[1]  # The minimum y value of B
    B_max_y = obj_b[3]  # The maximum y value of B
    
    # The maximum x value of A
    min_distance = int(table_width * 0.01)
    max_distance = int(table_width * 0.06)
    if A_max_x < B_min_x:
        # Calculate the X-axis distance between the two
        distance = B_min_x - A_max_x 
        # Calculate the overlapping area
        
        ...
        
        return score * overlap_ratio
    else:
        return 0  \# If A is not on the left of B, return 0 points
\end{verbatim}
}
</func>\\
Now you need to generate spatial functions based on the information I provided, including the relationship name, relationship definition, and scene spatial description of the object group that uses the relationship in the scenario.\\
You said the output function must be marked by </func>\\\\
Now begin to complete your mission:\\
- relationship name: <relationship\_name>\\
- relationship definition: <relationship\_definition>\\
- spatial description: <spatial\_description>\\
- relationship function:
\end{tcolorbox}

Similarly, we also construct the validation functions in the same manner. These validation functions take the object coordinates and their relationship as inputs and output a boolean result to indicate whether the relationship is consistent with the coordinates.

\subsection{Description Generation}

\begin{tcolorbox}[
    colback=gray!10,      
    colframe=gray!10,       
    width=\textwidth,     
    boxrule=0.5pt,        
    arc=1mm,              
    left=2mm, right=2mm,  
    top=2mm, bottom=2mm,   
]
You're a spatially imaginative assistant, and you can imagine what the expected scenario should look like based on the list of objects I've provided and the task instructions.
According to the 
relational library I provided, you'll describe the scenario in rigorous language that conforms to the physical rules of the real world, to human preferences, and to the requirements of the task instructions.
\end{tcolorbox}

\begin{tcolorbox}[
    colback=gray!10,      
    colframe=gray!10,       
    width=\textwidth,     
    boxrule=0.5pt,        
    arc=1mm,              
    left=2mm, right=2mm,  
    top=2mm, bottom=2mm,   
]    
**Definition:**\\
- Object list: In your description, the names of these objects must be consistent with those in the list. And when there are multiple examples of an object, they are distinguished in the list. That is, the object list: ["cup-0", "cup-1", "cup saucer"]. The object name in the description should also be the same as that in the list without "cup saucer-0".
\\- Relation Library: A well-defined relation contains its name and description, and you need to understand the meaning and use it. The relations can be shown as follow: <relations\_library\_key>.
- Relations between objects: You need to describe the relative relation of the appropriate combination of objects, including their relative position, alignment, and placement. You have to consider whether it's common sense to put them like this, and you also have to consider their size. In addition, to avoid ambiguity, you need to specify the axis of the relative relation. For example:\\
    1. The knives are placed on the right side of the spoon (*x* axis), with aligned at x-axis.\\
    2. The pencil is placed upright on one side of the book (*x* axis) with the tip facing the user, and they are align at the x-axis.\\
    
    3. Pencils and pens are placed on the top of the notebook (*z* axis), while the pancils and the pen are placed x-axis aligned.\\
    
    4. The cup saucer is on the top right of the plate (*xy* axis), with the cup on the top (*z* axis).\\
    
    5. Knives, forks, spoons they are all placed close together (*x* axis) and placed horizontally under the plate (*y* axis) with the knife is aligned with plate at y-axis.\\
\\\\
- All objects should be described at least once to ensure that they can be represented in a unique layout.
\\\\
**Step-by-step guidance:**\\
    1. You will first be provided with a list of objects and a task instruction. According to the instructions, you should describe the overall layout of the scene and any salient features, such as the instructions indicating that the user is left-handed, and you should focus on which objects should be placed on the left accordingly.
    
    2. You will identify semantic relations between objects. You should consider the functional, semantic, and geometric relations between objects.
    
    3. You will then sort the semantic asset groups according to the order in which they should be placed in the scene. You should consider the meaning of each group and the logic flow of the scenario layout. For example, you might place larger or more prominent assets first to establish the scene focus, such as a plate at a dining table or a notebook at a desk.
    
    4. Finally, you will format the grouping information into a clear and organized structure that is easy to understand by other designers or stakeholders. The output must be between the token </Description>.
\\\\
**Example:**\\
- Object List: ['coffee cup', 'mug', 'name plate', 'notebook', 'pen', 'pencil', 'water bottle']\\
- Task Instruction: Please set up the office table
\\\\
\#\#\# Step 1: Analyze the Task\\
    The task is to arrange the provided objects on an **office table**. The layout should be functional, logical, and ergonomic according to how a typical office table is used. Considerations include:
    \\
    ...
    \\
\\
\#\#\# Step 2: Identify Semantic Relations\\
    1. **Notebook and writing tools (pen and pencil)**: These are functionally related and should be grouped together for easy access during work. The pen and pencil should be positioned in a way that is convenient for writing.\\
    \\...
\end{tcolorbox}

\begin{tcolorbox}[
    colback=gray!10,      
    colframe=gray!10,       
    width=\textwidth,     
    boxrule=0.5pt,        
    arc=1mm,              
    left=2mm, right=2mm,  
    top=2mm, bottom=2mm,   
]    
\#\#\# Step 3: Organize Groups for Placement
    The objects will be grouped and placed in a logical order:\\
    1. **Primary workspace**: Notebook, pen, and pencil.\\\\
    ...
    \\\\
\#\#\# Step 4: Object Placement Description\\
</Description>\\
    \#\#\#\# Primary Workspace\\
        1. **Notebook**: Placed at the center of the table (*xy* axis), with its bottom edge parallel to the table's front edge.\\\\
        ...
        \\\\
</Description>\\\\

**Task:**\\
Now, I'm going to provide a sample for reference, and you'll need to think step-by-step based on the object list and task instructions and output a description of the prefetch scenario.\\
NOTE: it is very important to include all the object!!! And please do not change the name of the object.
\\\\
NOTE: Only volume alignment is considered, not edge alignment!!\\
\\
- Object List: <obj\_list>\\
- Task Instruction: <task\_instruction>\\
- Output:
\end{tcolorbox}

Based on the object list, task instructions, and available relations, we need to systematically generate a comprehensive and appropriate description to represent the expected scene. Then, we proceed to submit this response from the LLM to another LLM for validation, considering whether there are any deficiencies. Below is the prompt used for validation.

\begin{tcolorbox}[
    colback=gray!10,      
    colframe=gray!10,       
    width=\textwidth,     
    boxrule=0.5pt,        
    arc=1mm,              
    left=2mm, right=2mm,  
    top=2mm, bottom=2mm,   
]
You are a spatially imaginative and very rigorous layout evaluation assistant. You'll analyze the list of objects, task instructions, and corresponding scenario 
descriptions I've provided, and 
determine whether they meet the requirements of the task instructions and whether each object is properly placed.\\
**Step-by-Step Guide:**\\
    1. First, you will be provided with a list of objects, task instructions, and scenario description. Based on this information, you need to evaluate whether the scenario description is comprehensive and appropriate.\\
    2. **If there is a problem**, you need to list the deficiencies in the scenario description. The information must be in the tag </issue>.\\
    3. Finally, you only need to provide a Bool output, True or False, which must be in the tag </output>.\\\\
**Example:**\\
- Object List: ['coffee cup', 'mug', 'name plate', 'notebook', 'pen', 'pencil', 'water bottle']\\
- Task Instruction: Please set up the office table\\
- Description: \\\#\#\#\# Primary Workspace\\
        ...
\end{tcolorbox}
\begin{tcolorbox}[
    colback=gray!10,      
    colframe=gray!10,       
    width=\textwidth,     
    boxrule=0.5pt,        
    arc=1mm,              
    left=2mm, right=2mm,  
    top=2mm, bottom=2mm,   
]
\#\#\# Step 1 Analysis:\\
    The task instruction, "Please set up the office table," aims to organize the given objects in a structured and logical manner. Below is a step-by-step evaluation of the provided scenario description against the object list and task instruction:\\
\\
    1. **Object Coverage**:\\
    ...
\\
    2. **Spatial Clarity**:\\
    ...
\\
    3. **Comprehensiveness**: \\
    ...
\\

\#\#\# Step 2 Deficiencies (if any):\\
    The description is comprehensive and aligns well with the task instruction. However, there are **minor issues**:\\
    1. **Missing Positional Details**:\\
    - The `notebook` is placed at the center of the table, but there is no explicit mention of whether it is centered along both the *x* and *y* axes or just the *x-axis*. This could lead to slight ambiguity in placement.
    2. **Overlapping Concerns**:\\
    - The `name plate` and `notebook` share the center position on the table. It is unclear whether these objects overlap or whether the `name plate` is placed in front of the `notebook`.\\
\\
    Final Assessment:\\
    - The description is **mostly appropriate** and satisfies the task instruction, but the minor issues listed above could lead to slight misinterpretations.\\
\\
\#\#\# Output:\\

    </issue>
    1. Missing clarification on whether the `notebook` is centered along both *x* and *y* axes.
    2. Potential overlap between the `name plate` and `notebook` at the center of the table.
    </issue>\\\\

    </output>
    False
    </output>
\\
**Task:**\\
Now, I'm going to provide a sample for reference, and you'll need to think step-by-step based on the object list and task instructions and output a description of the prefetch scenario.\\
NOTE: it is very important to include all the object!!! And please do not change the name of the object.\\
NOTE: Only volume alignment is considered, not edge alignment!!\\
\\
- Object List: <obj\_list>\\
- Task Instruction: <task\_instruction>\\
- Description: <description>\\
- Output:
\end{tcolorbox}

When the system determines that the current description has no issues,it can then be handed over to the fast system to further generate the required discrete coordinate set and topological relations.

\subsection{Discrete Coordinate Generation}

Based on the generated scene description,we can use the following prompt to generate a discrete set of coordinates, which do not take into account the sizes of the objects but only consider their positional relationships.

\begin{tcolorbox}[
    colback=gray!10,      
    colframe=gray!10,       
    width=\textwidth,     
    boxrule=0.5pt,        
    arc=1mm,              
    left=2mm, right=2mm,  
    top=2mm, bottom=2mm   
]
You are a spatial imagination and logical layout generation assistant.
I want to create an arrangement of objects that is both ordered (physically feasible and visually appealing) and functional (based on instructions that meet the user's intended purpose). Now that I have a description of the spatial relationships of each object in the current scene, I need your help to further generate a set of **coarse-grained 3d poses** for a set of objects based on these descriptions.\\\\
**Definition:**\\
    - Scenario description: We will provide the placement of each object in the current scenario, which is the only basis for you to generate the relationship between each object.\\
    - 3D Pose: This is a right-hand coordinate system with the anchor object as the origin, and the other objects are placed in order. The format is [x,y,z]. These coordinates do not take into account dimensions, only their object location.\\
**Step-by-step guidance:**\\
    1. First, select an appropriate anchor object as the coordinate system origin based on the scene description. It is generally located in the middle of the layout. Like the dishes at the dining table, the notebook at the desk.\\
    2. Then place the objects according to their relative relationships, which you need to consider according to the axes they specify. If the plate is at [0,0,0] and the spoon is on its right (x-axis), the coordinates of the spoon are [1,0,0]. If an object is between them, you can put it directly between them, and the coordinate value is their average. For example, the knife is between the plate and the spoon, and the spoon is [0.5, 0, 0]\\
    3. Finally, format all 3D poses and output them to token</pose>.\\
\\
**Example:**\\
    - object list: ['coffee cup', 'mug', 'name plate', 'notebook', 'pen', 'pencil', 'water bottle']\\
    - Scene Description: \\\#\#\#\# Primary Workspace\\
        ...
        \\\\
    - Output: \\
        Here is the step-by-step process for generating **coarse-grained 3D poses** for the given object arrangement:\\
        \#\#\# **Step 1: Choosing the anchor object**\\
            - The **notebook** is the central object in the layout as mentioned in the description ("Placed at the center of the table"). It will serve as the origin of the coordinate system: **[0, 0, 0]**.\\
\\
        \#\#\# **Step 2: Calculating 3D poses**\\
\\
            \#\#\#\# **Primary Workspace**\\
                1. **Notebook**: \\
                - As the anchor object, the notebook is at the origin.  \\
                - **Pose**: `[0, 0, 0]`.
\\
 ...
\\\\
            \#\#\# **Step 3: Final formatted 3D poses**\\
                </pose>\\
                notebook: [0, 0, 0]\\
                pen: [1, 0, 0]\\
                pencil: [-1, 0, 0]\\
                ...\\
                </pose>\\
**NOTE: **\\
    it is very important to include all the object!!! And please do not change the name of the object.\\
\\
**Task: **\\
    Now, follow the step by step guidance finish the task.\\
\\
    - Object List: <object\_list>\\
    - Scene Description: <scene\_description>\\
    - Output:
\end{tcolorbox}

\subsection{Topological Relations Generation}

Similarly, while generating the discrete coordinate set,we also generate a set of topological relationships based on the scene description.

\begin{tcolorbox}[
    colback=gray!10,      
    colframe=gray!10,       
    width=\textwidth,     
    boxrule=0.5pt,        
    arc=1mm,              
    left=2mm, right=2mm,  
    top=2mm, bottom=2mm   
]
You are a spatially imaginative and logical layout generation assistant.
I want to create an arrangement of objects that is both ordered (physically feasible and visually appealing) and functional (based on instructions that meet the user's intended purpose). Now that I have a description of the spatial relationships of each object in the current scene, I need your help to further generate a set of relative relationships for a group of objects *based on these descriptions*.\\\\
**Definition:**\\
    - Scenario description: We will provide the placement of each object in the current scenario, which is the only basis for you to generate the relationship between each object.\\
    - Relative relationships: We provide a library of relative relationships, which are usually used to describe the relative positional relationships between objects. Format: relationship(ObjA, ObjB), where ObjB is the reference object\\
    - Relative position coordinate (RPC): The relative relationship we provide contains a list called RPC, which represents the relative coordinates between two objects. That is, (ObjA, ObjB)=[x,y,z] represents the coordinates [x, y, z] of ObjB in the right-hand coordinate system with ObjA as the origin.\\
\\
**Relationship Library:**\\
    <relationship\_library>\\
\\
**Step-by-step guidance:**\\
    1. First, select the appropriate relationship from the spatial relationship library according to the sequence in the scene description, and convert the description of each object into the spatial relationship related to the object. Crucially, you need to determine whether these descriptions and relationships are on the corresponding axis based on the RPCs in the relational library, rather than simply converting them based on the name of the relationship. For example, if the description is A is on the top of B (*y* axis), you should determine that the relationship described by the spatial relationship is above\_of in the relational database rather than on\_the\_top\_of according to the definition of RPC: [0,1,0].\\
    2. After you get these relationships initially, you need to further refine the relationships of these objects, because some objects are not fully described. For example, the coffee cup, mug and cup are all in the upper right corner of the plate, but it does not explain the relationship between the three cups. So you need to further refine their relationship, such as the mug on the left of the coffee cup, the cup on the right of the coffee cup, so add [left\_of('mug',coffee cup), right\_of(cup, coffee cup)]\\
    3. After completing these spatial relationships, you need to further consider how to adjust their alignment. You need to consider which objects should be aligned, in which direction, center or top or bottom, and so on. These alignments must also be looked up from the relational library.\\
    4. Finally, format all spatial relationships, *merge or delete* the extra spatial relationships, and output them to the token</relationships>.\\
\\
**Example:**\\
    - object list: ['coffee cup', 'mug', 'name plate', 'notebook', 'pen', 'pencil', 'water bottle']\\
    - Scene Description: \\\#\#\#\# Primary Workspace\\
        ...
\\\\
    - Output: \\
        \#\#\# Step 1: Initial relationships from scene description\\
            \#\#\#\# Primary Workspace:\\
\end{tcolorbox}
\begin{tcolorbox}[
    colback=gray!10,      
    colframe=gray!10,       
    width=\textwidth,     
    boxrule=0.5pt,        
    arc=1mm,              
    left=2mm, right=2mm,  
    top=2mm, bottom=2mm   
]
                1. **Notebook**: \\
                - `central\_column('notebook')` (Notebook is at the center of the table).\\
                - `near\_front\_edge('notebook')` (Bottom edge of notebook is near the table’s front edge).\\
\\...
\\
        \#\#\# **Step 2: Refinement of relationships**\\
            From the description, there are some implicit relationships that need to be added for completeness:\\
\\
                1. The relative relationship of missing name plate
                the relationship between name plate and notebook is above or below (Placed at the center of the table (*x* axis), facing outward from the user, with its bottom edge aligned to the table's edge.)\\
                However `near\_front\_edge('notebook')` is mentioned above, the object at the front edge of the desktop is a notebook, and the name plate is above the notebook (*y* axis).\\
                - `above\_of('name plate', 'notebook')`\\
\\
        \#\#\# **Step 3: Alignment considerations**\\
\\
            1. **Pen, Pencil, and Notebook**:\\
            - `aligned\_in\_x\_axis('pencil', 'notebook', 'pen')` (All three objects are horizontally aligned along the top edge of the notebook).\\
            \\
            2. **Coffee cup, Mug, and Water bottle**:\\
            - `aligned\_in\_x\_axis('mug', 'coffee cup', 'water bottle')` (All three are aligned along the table’s top edge).\\
\\
            3. **Name plate, Notebook**:\\
            - `aligned\_in\_y\_axis('name plate', 'notebook')` (Their bottom edge aligned to the table's edge).
\\\\
        \#\#\# **Step 4: Final formatted relationships**\\
            Here is the complete set of spatial relationships formatted into tokens:\\
            </relationships>\\
            central\_column('notebook')\\
            near\_front\_edge('notebook')\\
            right\_of('pen', 'notebook')\\
            left\_of('pencil', 'notebook')\\
            aligned\_in\_x\_axis('pencil', 'notebook', 'pen')\\
            right\_above\_of('coffee cup', 'notebook')\\
            left\_of('mug', 'coffee cup')\\
            right\_of('water bottle', 'coffee cup')\\
            aligned\_in\_x\_axis('coffee cup', 'water bottle', 'mug')\\
            central\_column('name plate')\\
            above\_of('name plate', 'notebook')\\
            aligned\_in\_y\_axis('name plate', 'notebook')\\
            </relationships>\\
\\\
**NOTE: **\\
    it is very important to include all the object!!! And please do not change the name of the object. Cannot generate objects outside the list!!!\\
    The spatial relationship must be used according to the definition in the relational library. For example,'central\_column' is a "Unary" relationship, and the object described is one.\\
    The alignment considered should be the alignment of the volume of the object, not the edge alignment. Object groups placed along the x axis cannot be aligned on the y axis. Similarly, object groups placed along the y or z axis can only be aligned on the y or z axis.\\
    The relationship to be used must be in the library. Relationships that are not defined in the library cannot be used.\\
\\
**Task: **\\
    Now, follow the step by step guidance finish the task.\\
\\
    - Object List: <object\_list>\\
    - Scene Description: <scene\_description>\\
    - Output:
\end{tcolorbox}

\clearpage
{
    \small
    \bibliographystyle{unsrtnat}
    \bibliography{ref}
}



\end{document}